\documentclass[letterpaper, 10 pt, conference]{ieeeconf}  

\IEEEoverridecommandlockouts                              

\usepackage{scrextend}
\usepackage{pslatex}
\usepackage{graphicx}
\usepackage{float}
\usepackage[hyphens]{url}
\usepackage[ruled,vlined]{algorithm2e}
\PassOptionsToPackage{hyphens}{url}\usepackage{hyperref}
\title{\LARGE \bf
Autonomous Vehicle Benchmarking using Unbiased Metrics
}

\author{David Paz$^{1*}$, Po-jung Lai$^{1*}$, Nathan Chan$^{2*}$, Yuqing Jiang$^{2*}$ Henrik I. Christensen$^{2*}$
\thanks{*This work was performed in collaboration with UC San Diego's Operations, Mailing Center, and Police Station.}
\thanks{$^{1*}$Department of Electrical and Computer Engineering,
        University of California, San Diego, 9500 Gilman Dr, La Jolla, CA 92093}
\thanks{$^{2*}$Department of Computer Science and Engineering,
        University of California, San Diego, 9500 Gilman Dr, La Jolla, CA 92093}%
}

\begin{document}

\maketitle
\thispagestyle{empty}
\pagestyle{empty}

\begin{abstract}

With the recent development of autonomous vehicle technology, there have been active efforts on the deployment of this technology at different scales that include urban and highway driving. While many of the prototypes showcased have been shown to operate under specific cases, little effort has been made to better understand their shortcomings and generalizability to new areas. Distance, uptime and number of manual disengagements performed during autonomous driving provide a high-level idea on the performance of an autonomous system but without proper data normalization, testing location information, and the number of vehicles involved in testing, the disengagement reports alone do not fully encompass system performance and robustness. Thus, in this study a complete set of metrics are applied for benchmarking autonomous vehicle systems in a variety of scenarios that can be extended for comparison with human drivers and other autonomous vehicle systems. These metrics have been used to benchmark UC San Diego’s autonomous vehicle platforms during early deployments for micro-transit and autonomous mail delivery applications.

\end{abstract}

\section{INTRODUCTION}

Autonomous vehicle technology has been under active development for at least 30 years \cite{c1} \cite{c2} \cite{c3} \cite{c4}. Since the time the technology was first conceived \cite{c5}, a wide range of applications have been explored from micro-transit to highway driving applications but more recently has started to become commercialized. With the variety of use cases in question, one important topic involves safety. This has received the attention of state officials, and in many cases, regulations and policies have been imposed.

In some states, the Department of Motor Vehicles requires a summary of disengagement reports from each entity performing tests on public roads to provide a better understanding on the number of annual interventions each self-driving car entity is generating. In the state of California alone, the Department of Motor Vehicles (DMV) requires autonomous vehicle companies with a valid testing permit to submit annual reports with a summary of system disengagements. At the time this paper is being written, 66 tech entities hold a valid autonomous vehicle testing permit and only three hold a driverless testing permit.%
\footnote{\url{https://www.dmv.ca.gov/portal/dmv/detail/vr/autonomous/permit}} 

Even though many of these reports include certain information to estimate the number of disengagements performed in an entire year, most of the publicly available disengagement reports%
\footnote{\url{https://www.dmv.ca.gov/portal/dmv/detail/vr/autonomous/disengagement_report_2019}} 
are not time and distance normalized: did the vehicle experience five disengagements during the course of 10 miles or 10,000 miles? Or did it experience five disengagements over the course of 10 minutes or 2,000 hours?

Given the lack of spatiotemporal information, in many cases, these unnormalized reports make it impossible to quantify the performance and robustness of the autonomous systems and most importantly quantify their overall safety with respect to other autonomous systems or human drivers. 

This study aims to shed light on autonomous system technology performance and safety by leveraging spatiotemporal information and metrics geared towards benchmarking Level 3 to Level 5 autonomous vehicle systems.\footnote{\url{https://www.nhtsa.gov/technology-innovation/automated-vehicles-safety}} Our key contributions consist of three different parts:
\begin{itemize}
   \item We introduce the concept of intervention maps for disengagement visualization and analysis. Additionally, the metrics we introduced in \cite{c6} have been extended in order to account for safety driver dependability.
   \item With spatial information as a function of time, we separate the results into different road types to provide more realistic and objective comparisons across different platforms without biasing\footnote{We define \textit{unbiased} in the context of making objective comparisons across different vehicle platforms without biasing towards a specific system.} the results.
   \item A four-month data collection phase is performed using UCSD's Autonomous Vehicle Laboraty autonomous vehicles; the data is analyzed using the metrics proposed.
 \end{itemize}
 With the methods introduced in this study, our team plans on open sourcing an online tool for autonomous vehicle benchmarking to encourage autonomous vehicle entities to report their data in order to objectively quantify system safety and long term autonomy capabilities. 

\section{Related Work}
The areas of autonomous vehicle benchmarking have remained relatively unexplored. Prior related work in the area of benchmarking sheds light on performance measures for intelligent systems in off-road and on-road unmanned military applications\cite{c9}. While the performance measures proposed may serve for certain unmanned military applications, autonomous vehicle applications in public road conditions often require safety drivers to ensure the vehicles will not behave erratically and pose danger for road users if failure cases arise. 

With road user safety and failure cases in mind, \cite{c10} focuses on estimating the number of miles a self-driving vehicle would have to be driven autonomously in order to demonstrate its reliability with respect to human drivers and proof of their safety. This study specifically shows that self-driving vehicles will take tens to hundreds of years to demonstrate considerable reliability over human drivers with respect to fatalities and injuries. In addition, this naturally leads to the questions, how can the autonomous vehicle progress in between be measured objectively? 

While certain self-driving car entities have identified the flaws with current disengagement data reported by the DMV \cite{c11} \cite{c12} \cite{c13}, to the best of our knowledge, our team is the first to make objective comparisons of autonomous systems by studying their long term autonomy implications using real autonomous vehicle data collected from diverse and realistic urban scenarios.

\section{Metrics}
In this section, the metrics and tools used to benchmark an autonomous vehicle during a four-month study at UC San Diego are defined with the goal of fully characterizing the performance of the systems over time. 

\subsection{Direct System Robustness Characterization}

For direct system robustness characterization, the metrics of choice are given by Mean Distance Between Interventions (MDBI) and Mean Time Between Interventions (MTBI). These metrics provide a normalized means of benchmarking system robustness over time by including temporal and spatial information. This makes them ideal for comparing performance against other systems. In contrast, unnormalized data cannot be used to perform objective comparisons across different autonomous vehicle systems as one cannot estimate how often the disengagements are happening in terms of time and distance. For this reason, we do not use intervention counts alone to quantify performance. By definition, the MDBI and MTBI statistics can be computed as shown in Equation \ref{equation:mdbi} and \ref{equation:mtbi}.

\begin{equation}
\mathrm{MDBI}=\frac{\mathrm{ Total\ Distance }}{\mathrm{ Number\ of\ Interventions }}
\label{equation:mdbi}
\end{equation}

\begin{equation}
\mathrm{MTBI}=\frac{\mathrm{ Total\ Uptime }}{\mathrm{ Number\ of\ Interventions }}
\label{equation:mtbi}
\end{equation}

\begin{figure}[H]
\centering
\includegraphics[width=0.4\textwidth]{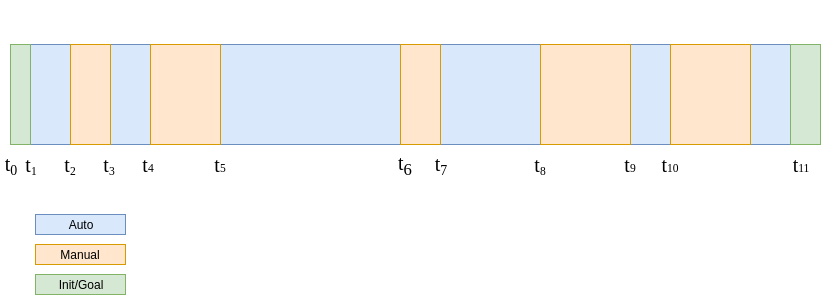}
\caption{ Enable/disable (disengagement) signal as a function of time. }
\label{figure:interventions}
\end{figure}

While the definitions for MDBI and MTBI are direct, the measurements for distance, uptime and the number of interventions require the data to be separated into two different categories: the first corresponds to the time elapsed and distance traveled in autonomous mode and the second to the time elapsed and distance traveled during manual driving. By separating these into two different sets of data, the effective system robustness can be measured in regards to its dependence on a safety driver if one or more manual interventions are performed. Nevertheless, in order to separate the manual and autonomous data, vehicle disengagement information must be recorded as a function of time as close to real-time as possible; this can be visualized in Figure \ref{figure:interventions}.

In the figure, manual and autonomous driving segments are represented by orange and blue colors, respectively, where the separation is given by an intervention or a system re-enable signal. Given that a manual intervention could be performed for an arbitrary length of time, it is important to accurately measure the disengagement signals in real-time by associating them with a system timestamp. While these measurements can be performed by manual annotation, this introduces human error. Therefore, in the measurements performed in this study, each autonomous vehicle was retrofitted with a logging device that records the enable and disable signals over time by using Unix time. This device operates in an encapsulated environment and records serialized data for vehicle pose, speed, enable signals, as well as their corresponding timestamp. Given this data, measuring the time elapsed between a disengagement and a re-enable signal can be measured by the difference in timestamps. On the other hand, two methods can be employed for measuring the distance traveled in between any two given timestamps $t_i$ and $t_{i+k}$, where $t_i < t_{i+k}$ as shown in Equation \ref{equation:distancedef1} and \ref{equation:distancedef2}--where the vehicle position at time $t$ is given by $\textbf{X}_t=[x_t,y_t,z_t]^\top$ and speed is given by $v_t$. For the measurements performed in this study, Equation \ref{equation:distancedef1} was used for estimating distance given that vehicle pose estimates are provided with a high degree of precision by the LiDAR based Normal-Distributions Transform localization algorithm \cite{c7}. The devices used for these measurements are also introduced in our previous work on the lessons learned from deploying autonomous vehicles \cite{c6} and a high level description will be provided in the next section.

\begin{equation}
\sum_{\tau=i+1}^{i+k} \left\|\textbf{X}_{\tau}-\textbf{X}_{\tau-1}\right\|
\label{equation:distancedef1}
\end{equation}

\begin{equation}
\sum_{\tau=i+1}^{i+k} v_{\tau}\left(t_{\tau}-t_{\tau-1}\right)
\label{equation:distancedef2}
\end{equation}

By measuring the distance covered by the ego-vehicle along with its associated uptime in between a disengagement and a re-enable signal (manual mode) or in between re-enable signal and a disengagement (autonomous mode), MDBI and MTBI can be extended to cover both, manual driving and autonomous driving as shown in Equations \ref{equation:mdbi_a}-\ref{equation:mtbi_m}

\begin{equation}
\mathrm{MDBI}_{A}=\frac{\mathrm {Total\ Auto\ Distance}}{\mathrm {Number\ of\ Interventions}}
\label{equation:mdbi_a}
\end{equation}

\begin{equation}
\mathrm{MTBI}_{A}=\frac{\mathrm {Total\ Auto\ Uptime}}{\mathrm {Number\ of\ Interventions}}
\label{equation:mtbi_a}
\end{equation}

\begin{equation}
\mathrm{MDBI}_{M}=\frac{\mathrm {Total\ Manual\ Distance}}{\mathrm {Number\ of\ Interventions}}
\label{equation:mdbi_m}
\end{equation}

\begin{equation}
\mathrm{MTBI}_{M}=\frac{\mathrm {Total\ Manual\ Uptime}}{\mathrm {Number\ of\ Interventions}}
\label{equation:mtbi_m}
\end{equation}

$\mathrm{MDBI}_{A}$, $\mathrm{MTBI}_{A}$, $\mathrm{MDBI}_{M}$ and $\mathrm{MTBI}_{M}$ measure the overall system robustness but also provide additional measures on how dependent the system is on a safety driver if any disengagements are performed: $\mathrm{MDBI}_A$ and $\mathrm{MTBI}_A$ measure the average distance and time an autonomous car is capable of driving without any interventions, while $\mathrm{MDBI_M}$ and $\mathrm{MTBI_M}$ measure the average manual input required by a safety driver in terms of distance and time elapsed in a mean sense. In an ideal system, the number of interventions is zero; we can address the limit forms as follows: $\mathrm{MDBI}_{A} = \frac{C}{0} \rightarrow \infty$, $\mathrm{MTBI}_{A} = \frac{C}{0} \rightarrow \infty$, $\mathrm{MDBI}_{M} = \frac{0}{0} \rightarrow 0$, and $\mathrm{MTBI}_{M} = \frac{0}{0} \rightarrow 0$.

\subsection{Intervention Maps}

Although with the metrics introduced, a statistical analysis can be performed across multiple vehicles with different autonomous systems, different road types can influence the metrics, i.e, did the vehicle drive on a testing track or did it engage in high-traffic scenarios? To incorporate the diverse environments an autonomous vehicle must navigate through into our benchmarking tools, we introduce the concept of intervention maps and separate the data by different road types. 

Intervention maps are specific to testing routes or geographical areas during benchmarking and are encoded in an occupancy grid format that contains normalized disengagement counts over time. This information can be extracted by associating disengagement information with spatial data as given in Algorithm \ref{alg:intervention_count} but also be useful for estimating MDBI and MTBI for a specific area or region. With this intervention occupancy map, the normalized values $[0,1]$ can be mapped to a color gradient for visualization purposes. Furthermore, by declaring a time and distance range for a particular location, these maps can help visualize disengagement patterns and also provide a sense on the quality of the data based on the location, MDBI and MTBI. This approach makes it a much more adequate benchmark across self-driving platforms. 

\begin{algorithm}[h]
\SetAlgoLined
\KwData{Enable/disable signal $\textbf{DBW}=[\textbf{DBW}_1, \textbf{DBW}_2, ..., \textbf{DBW}_n]$, Vehicle Pose $\textbf{X} = [\textbf{X}_1, \textbf{X}_2, ..., \textbf{X}_n]$
}
\KwResult{Normalized intervention occupancy grid $\textbf{M}$}

\CommentSty{\#Disengagement and pose association}

\textbf{DBWPose} = []

 \textbf{M} = [][]
 
 \For{$\textbf{X}_i \in \textbf{X}$ and $\textbf{DBW}_i \in \textbf{DBW}$}{
  \textbf{DBWPose}.append(($\textbf{X}_i$, $\textbf{DBW}_i$, closest\_timestamp))
 }
 \CommentSty{\#Populate occupancy grid}
 
 \For{$\textbf{DBWPose}_i \in \textbf{DBWPose}$}{
    \If{$\textbf{DBWPose}_i[1] == False$}{
        \textbf{M}[$\lfloor\textbf{DBWPose}_i[0][0]\rfloor$][$\lfloor(\textbf{DBWPose}_i[0][1]\rfloor$] $+=1$
    }
 }
 
 \CommentSty{\#Normalize}
 
 \textbf{M} $/=$ \textbf{$max$(\textbf{M})}
 
 \caption{Intervention Count and Normalization using an occupancy grid. }
 \label{alg:intervention_count}
\end{algorithm}

\subsubsection{Enhancing Intervention Map Representation} 
While in the Results sections, a number of patterns are identified based on the observations from intervention maps, to provide additional context to the information that is being visualized, additional road network information can be incorporated.

In this case, we separate every trip into individual road segments depending on a set of predefined conditions: (1) Dynamic Road, (2) Regular Road, (3) Freeway, and (4) Development/Private. A dynamic road corresponds to road segments without explicit lane definitions that include dynamic interactions with other road users such as alleys and pedestrian walkways. Regular roads on the other hand correspond to well-defined and roads with speed limits and fully defined right-of-way rules. Freeway road segments correspond to roads with continuous lane definitions and no intersections. Development or private roads correspond to testing-and-evaluation road segments that are well-controlled for system development whereas (1)-(3) correspond to realistic and uncontrolled environments. Lastly, each road segment is associated with a speed limit.

\begin{figure}[H]
\centering
\includegraphics[width=0.25\textwidth]{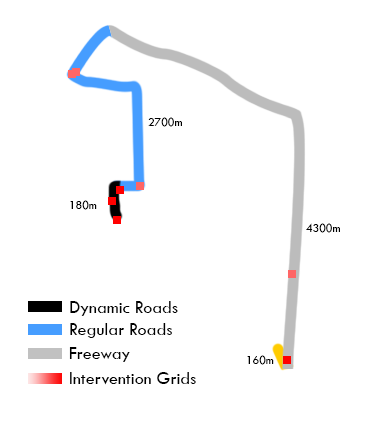}
\caption{ Sample intervention map with different route types. }
\label{figure:sample_map}
\end{figure}

A sample occupancy grid map with arbitrary road definitions and types can be seen in Figure \ref{figure:sample_map}. With the distance of each road segment and the class types, each trip or planned mission can incorporate road information in terms of a percentage of the total distance traveled. For instance, this particular trip corresponds to 2.4\% dynamic roads, 39.0\% regular roads, and 58.6\% freeway road segments. With this type of data separation, one can define a sequence of benchmarks between one or more autonomous vehicle platforms under the same road types, distances and their associated MTBI and MDBI metrics.

\subsection{Autonomous vs Manual Driving Benchmarking}
An extension to $\mathrm{MDBI}_A$, $\mathrm{MTBI}_A$, $\mathrm{MDBI}_M$, and $\mathrm{MTBI}_M$, involves human driver to autonomous system comparison. Table \ref{table:auto_manual_metrics} corresponds to an additional set of metrics introduced in our previous work \cite{c6} that can further explain the differences between human drivers and autonomous vehicles in terms of energy consumption, maintenance cost, and control. For example, depending on the steering, acceleration and braking control inputs, more energy may be required to drive along the same routes if a system overcompensates for small errors. As a result, this can impact energy consumption, brake and tire wear. These cumulative effects can affect the overall cost of ownership of a vehicle, as well as the environmental impact. For benchmarking purposes, the measured steering, acceleration and braking status reports can be compared in the frequency domain for autonomous and manual driving.

While in the experiments section, human driver data is not included for direct comparison in terms of energy consumption or maintenance cost, these methods have been used for benchmarking level-4 autonomous trucks as part of a joint TuSimple/UC San Diego effort.\cite{c14}

\begin{table}[H]
\caption{Metrics for bench marking autonomous vs manual driving.}
\label{table:auto_manual_metrics}
\begin{center}
\begin{tabular}{|p{1.5cm}||p{4cm}||p{1.5cm}|}
\hline
\textbf{Trigger} & \textbf{Metric} & \textbf{Type}\\
\hline
Energy & Miles per Gallon (MPG) or Charged Consumed & Continuous \\
\hline
Maintenance Cost & Brakes and Tire Wear & Continuous \\
\hline
Up-time & Time Elapsed Per Trip & Event Driven \\
\hline
Control & Speed, Acceleration, Steering Angle Fourier Transform & Continuous \\
\hline
\end{tabular}
\end{center}
\end{table}

\section{DATA COLLECTION}
\begin{figure}[H]
\centering
\includegraphics[width=0.3\textwidth]{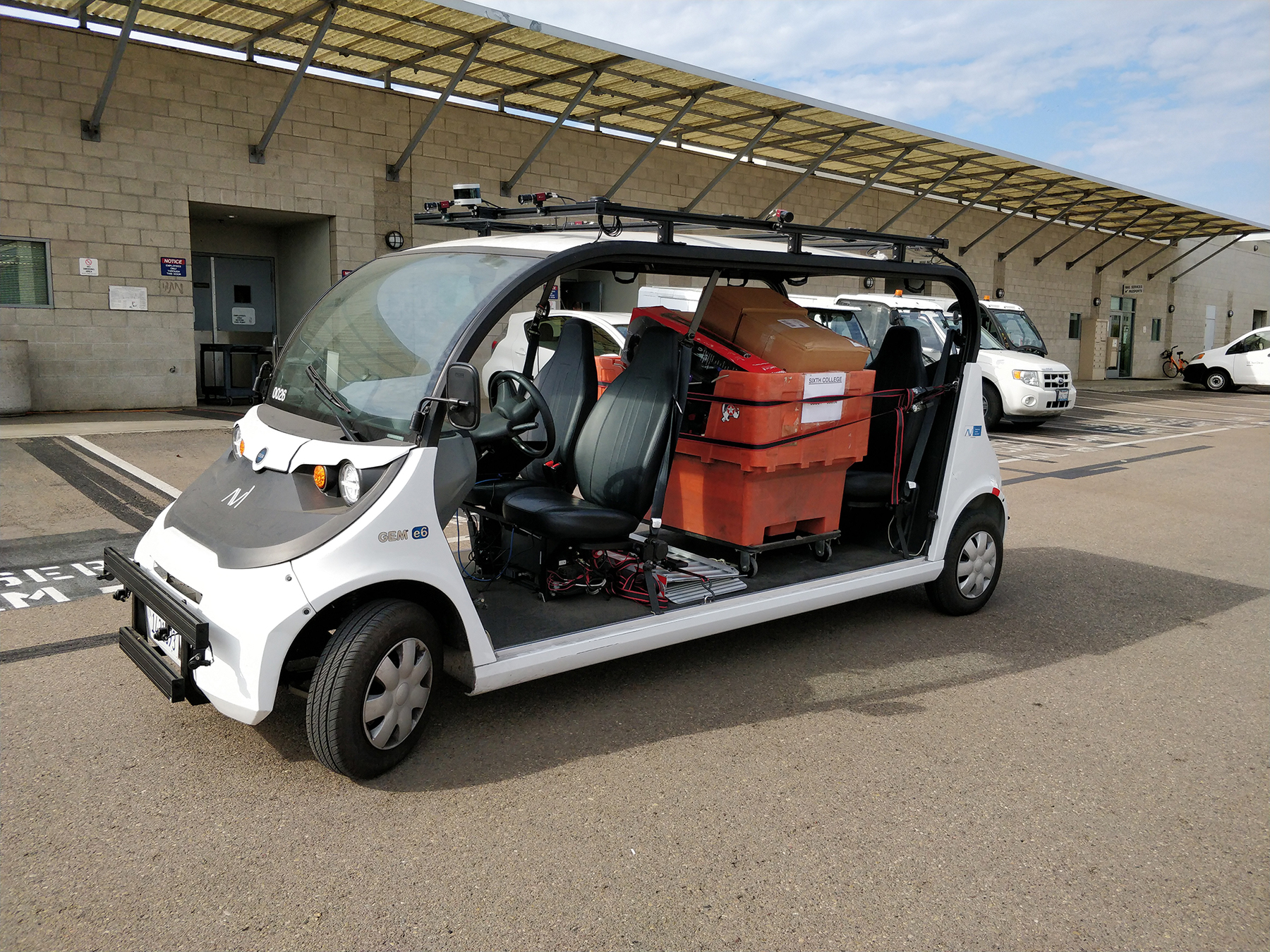}
\caption{UC San Diego's autonomous mail delivery vehicle carrying packages and mail. }
\label{figure:av_0026}
\end{figure}

As part of a collaborative effort between UC San Diego’s Autonomous Vehicle Laboratory (AVL), Mailing Center, Fleet Services, and Police Department, a GEM e6 electric vehicle (Figure \ref{figure:av_0026}) retrofitted with a complete drive-by-wire system and full sensor suites was used for conducting field tests at the UC San Diego campus. The design strategies and implementations used in the course of this study are described in \cite{c6}; however, the data and results presented in this paper are not part of it. 

\subsection{Vehicle Signals Recorded}

For the data collection process, our team worked closely with the mailing center to deploy the vehicles for autonomous mail delivery applications over Summer and Fall 2019 while continuously monitoring the systems and collecting data. The vehicle operated under highly dynamic and stochastic environments such as areas with high pedestrian and vehicle interactions, and construction sites. To record the various signals required for benchmarking, two tools were used as the basis for data logging: the ROSBAG format \cite{c8}, as well as a Raspberry Pi logging device that received serialized data and stored it in SQLite databases. Table \ref{table:vehicle_signals} corresponds to the different signals recorded as functions of epoch/Unix timestamps. It should be noted that for every autonomous mission, manual notes were taken to log the type of interventions performed, safety driver\footnote{During the data collection process, the same safety driver performed all of the manual interventions.} information, and the weather conditions. These notes are most useful for understanding bottlenecks and improving system performance.

\begin{table}[h]
\caption{Vehicle Signals Recorded.}
\label{table:vehicle_signals}
\begin{center}
\begin{tabular}{|p{3cm}||p{3.5cm}|}
\hline
\textbf{Signal} & \textbf{Representation} \\
\hline
Vehicle Pose (local map frame) & $\textbf{P}=[\textbf{X}^\top , \textbf{Q}^\top ]^\top$

$\textbf{X} = [x,y,z]^\top$ (meters) 

$\textbf{Q} = [q_0,q_1,q_2, q_3]^\top$ 

(quaternion)\\
\hline
GPS & Latitude

      Longitude
      
      Altitude (ft) \\
\hline
IMU & $\textbf{a} = [a_x, a_y, a_z]^\top$ (m/s$^2$)
    
$\textbf{w}=[w_x, w_y, w_z]^\top$ (s$^{-1}$)\\
\hline
Vehicle Speed & $v$ (m/s)  \\
\hline
Vehicle Target Speed & $v$ (m/s)  \\
\hline
Enable/Disable Signal & $0$ - Disabled

$1$ - Enabled\\
\hline
Acceleration & $[0,1]$ (Unitless)\\
\hline
Brake Control & $[0,1]$ (Unitless)\\
\hline
\end{tabular}
\end{center}
\end{table}

\subsection{Missions}
The autonomous mail delivery missions performed in this study consist of two routes within the UC San Diego campus: Warren College and Sixth College--where a trip or mission is defined to be as a round trip from the mailing center to the drop point and back. Round trip distances to Warren College and Sixth College correspond to 1,903m and 1,588m, respectively. In total, there are 24 trips to Warren College and 29 trips to Sixth College.

\begin{figure}[H]
\centering
\includegraphics[width=0.4\textwidth]{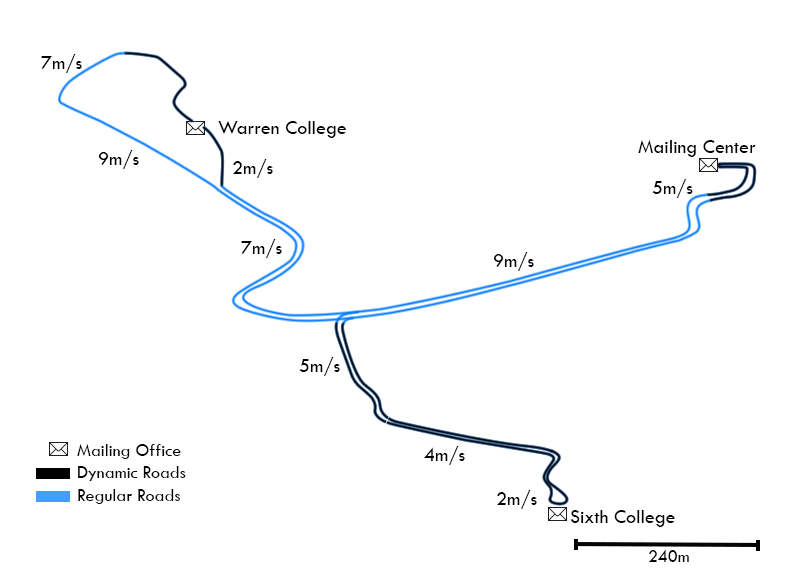}
\caption{Warren and Sixth colleges routes with road type information and speed limits. }
\label{figure:warren_sixth_base}
\end{figure}

For the intervention map representation of the areas covered, the different segments have been classified as either dynamic or regular roads. A map generated using the vehicle pose with the corresponding road types is represented in Figure \ref{figure:warren_sixth_base}, where a trip to Warren College consists of 412m of dynamic road segments and 1,492m of regular road segments. On the other hand, a trip to Sixth College consists of 916m of dynamic road segments and 672m of regular road segments. In other words, 21.6\% of Warren College trips correspond to dynamic road navigation and 57.7\% of Sixth College trips correspond to dynamic road navigation. 

For every trip performed, a trained safety driver was responsible for supervising the vehicle continuously. At the same time, a second team member recorded manual notes about trip information, intervention details, as well as monitored the system. 

\section{RESULTS}
\subsection{MTBI and MDBI Results}
Between the summer and fall 2019 mail delivery missions, the data from a single autonomous vehicle corresponds to 89.9km and 6.9 hours in autonomous mode.

The MTBI and MDBI statistics are represented in Table \ref{table:mtbi_mdbi_results} for summer and fall quarters respectively. The collective statistics from both quarters are shown in the third row.

\begin{table}[H]
\caption{MDBI (meters/intervention) and MTBI (seconds/intervention) intervention summary for summer and fall quarters}
\label{table:mtbi_mdbi_results}
\begin{center}
\begin{tabular}{|c||c||c||c||c|}
\hline
\textbf{Quarter} & $\textbf{MDBI}_A$ & $\textbf{MTBI}_A$ & $\textbf{MDBI}_M$ & $\textbf{MTBI}_M$ \\
\hline
Summer 2019 & 414.201 & 113.82& 24.0 & 12.77  \\
\hline
Fall 2019 & 283.08 & 84.44& 19.25 & 11.54  \\
\hline
Overall & 380.42 & 106.25& 22.77 & 12.46  \\
\hline
\end{tabular}
\end{center}
\end{table}

From the $\mathrm{MDBI}_A$ and $\mathrm{MTBI}_A$ metrics in Table \ref{table:mtbi_mdbi_results}, one can infer that, on average, the vehicle drove autonomously 380m or for 106 seconds before an intervention was made. In terms of the safety driver dependability that the $\mathrm{MDBI}_M$ and $\mathrm{MTBI}_M$ metrics model, on average, the safety driver intervened for 22.77m or for 12.46 seconds. Furthermore, it can be observed that the statistics significantly vary between summer and fall quarters.
This significant difference can be explained by campus traffic and ongoing activities experienced
early in fall quarter. In the fall, the mail delivery routes experience higher traffic and foot
activity from students moving in or starting classes. Separating these results based on time and testing location can help explain trends and traffic patterns. At the same time, it
is important to estimate collective averages to make note of the impact of the software release
versions on the overall robustness. 

\subsection{Intevention Map}
\begin{figure}[htb]
\begin{minipage}[t]{.23\textwidth}
\centering
\includegraphics[width=\textwidth]{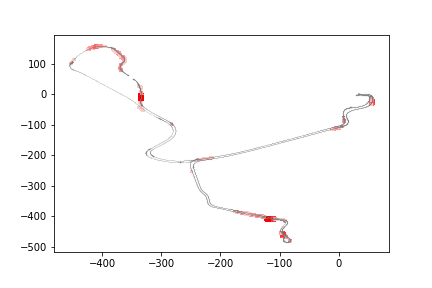}
\end{minipage}
\hfill
\begin{minipage}[t]{.23\textwidth}
\centering
\includegraphics[width=\textwidth]{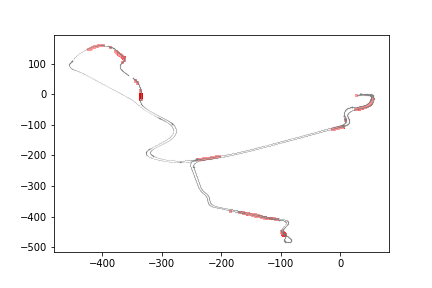}
\end{minipage}  
 
\caption{Automatically generated intervention maps for summer (left) and fall (right) 2019 quarters. Units in meters.}
\label{figure:auto_int}
\end{figure}

\begin{figure}[H]
\centering
\includegraphics[width=0.4\textwidth]{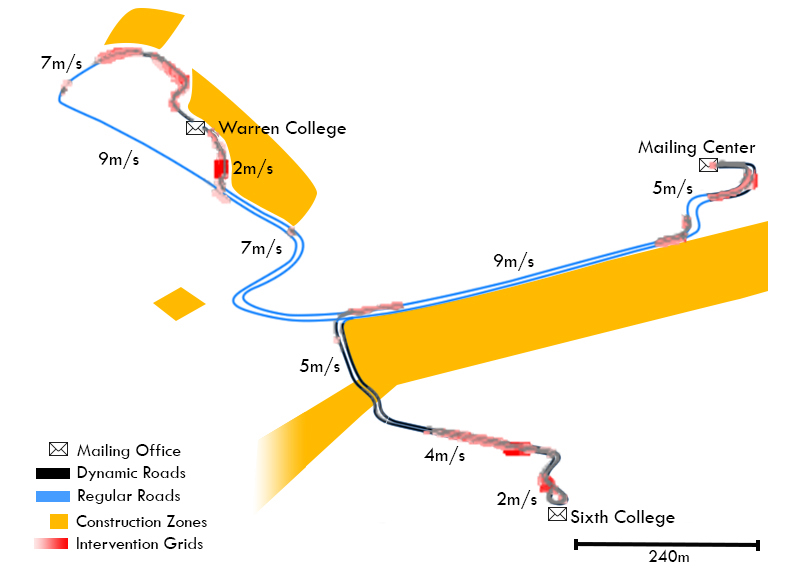}
\caption{
Overall intervention map
}
\label{figure:fall_summer_int_map}
\end{figure}

By applying the intervention map tools introduced, the automatically generated occupancy grid map with raw intervention count data can be seen in Figure \ref{figure:auto_int}. For better visualization, these maps have been re-scaled and superimposed as shown in Figure \ref{figure:fall_summer_int_map}. This figure corresponds to the aggregate data from summer and fall quarters and includes construction zones to better understand the campus dynamics.

In general, the areas with higher interventions occur around intersections but also along dynamic environments and construction sites. Without including this information, it is not straightforward to identify short-comings while processing large collections of data. More specifically, the Warren College mailing center path corresponds to a fork between a wide pedestrian walkway and the main road that is used during mail delivery. While our autonomous vehicle is permitted to drive along those areas while enforcing a 2m/s speed limit, the stochastic interactions with pedestrians are challenging.\footnote{By law, the campus speed limit is set to 25mph but in order to ensure safety along pedestrian-shared paths, the autonomous vehicle must adjust to different roads. Therefore, some roads require speed adjustments to be performed.} This same pedestrian walkway is protected by metallic bollards, requiring a manual intervention quite often since the spacing of the bollards leaves approximately 11cm of clearance on each side of the vehicle.

\begin{figure}[H]
\centering
\includegraphics[width=0.4\textwidth]{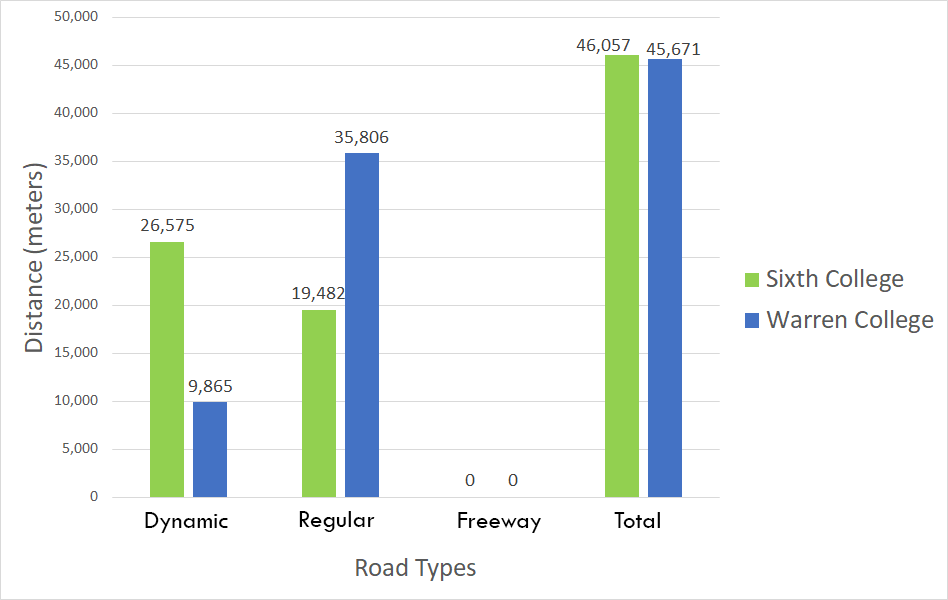}
\caption{Distances travelled for urban mail delivery routes: Warren and Sixth colleges. AVs did not engage in highway/freeway driving.}
\label{figure:distances_overall}
\end{figure}

As previously noted, the data collected from both quarters is a combination of regular and dynamic roads. Out of the 1,903m round-trip to Warren College, 411.96m correspond to dynamic road segments and 1,492m correspond to regular roads. On the other hand, out of the 1,588m round-trip to Sixth College, 916.2m correspond to dynamic roads and 671.98m to regular roads. In terms of the road categories considered here, Figure \ref{figure:distances_overall} allows us to visualize the variation and complexity on the types of roads in which the vertical axis corresponds to the distance for each road category. This illustrates the importance of the quality of the data being benchmarked: while the autonomous vehicle covered similar overall distances to each college, the variation between regular and dynamic roads is significant. Although our autonomous vehicle did not engage in highway/freeway driving, objective comparisons with respect to other systems can still be performed by defining separate benchmarks for each of the road types. In our case, our vehicles are not intended for highway driving.

\section{CONCLUSION AND FUTURE WORK}

With the autonomous vehicle data collected from mail delivery missions at UC San Diego during the an initial deployment phase, the overall vehicle performance has been quantified in terms of its capabilities to operate without assistance ($\mathrm{MTBI}_A$ and $\mathrm{MDBI}_A$), its dependability on human input ($\mathrm{MTBI}_M$ and $\mathrm{MDBI}_M$), by utilizing the concept of intervention maps, and the type of road conditions that are influenced by variation and the quality of the data. 
While, in a mean sense, the autonomous mail delivery vehicle required a safety driver intervention every 380m with an average human intervention lasting 23m, the techniques introduced in this study have provided a means of analyzing patterns from the mail delivery missions that are being actively used to address system shortcomings such as improvements on pedestrian and vehicle intent recognition and dynamic planning. 
To encourage other autonomous vehicle entities to benchmark their autonomous vehicle systems with the methods proposed, our team plans on open-sourcing the data collected from the mail delivery missions along with an online tool to objectively compute the overall system robustness as a function of the quality of the miles traversed, the georeferenced locations of the data collected, and the safety considerations introduced by \cite{c10}. We expect that the dissemination of these methods and tools will raise awareness on the overall performance of state-of-the-art autonomous vehicle technology in order to better understand the shortcomings of today’s technology and collectively design better performing systems.

\addtolength{\textheight}{-12cm}   




\section*{ACKNOWLEDGMENT}
We appreciate the support from Timothy Wheeler and Scott Driscoll from the UC San Diego Mailing Center, as well as Shiqi Tang and Andrew Liang from the AVL for assisting on multiple parts of this project and maintaining the vehicles. We are also grateful for the support we have received from campus operations, facilities and police station. Without their support, this project would not be possible.


\end{document}